\documentclass{article}

\usepackage[final]{corl_2026} 

\usepackage{wrapfig}
\usepackage{graphicx}
\usepackage{amsfonts}
\usepackage{amsmath}
\usepackage{amssymb}
\usepackage{booktabs}
\usepackage{multirow}
\usepackage{float}
\usepackage{kotex} 
\usepackage{rotfloat}

\usepackage{caption}
\usepackage{subcaption}
\usepackage{bbm}

\newcommand{\ms}[2]{$#1_{\pm #2}$}
\newcommand{\msb}[2]{$\mathbf{#1}_{\pm #2}$}

\usepackage{microtype}
\title{
CompVLA: A Variable Compliance Vision-Language- Action Model for Contact-rich Manipulation
}

\author{
  Jongmin Kim$^{*1}$ \: Junsu Ha$^{*1}$ \: Che-Sang Park$^{*1}$ \: Minchang Song$^{1}$ \: Hyeokju Jeong$^{1}$ \\
  \textbf{Himchan Hwang$^{1}$ \: Jianlong Fu$^{2}$ \: Frank C.~Park$^{1}$} \\
  $^{1}$Seoul National University, $^{2}$Microsoft Research Asia \\
  \texttt{\{jmkim, hajunsu, cspark, smc, hjjeong, himchan\}@robotics.snu.ac.kr, } \\
  \texttt{jianf@microsoft.com, fcp@snu.ac.kr}
}

\begin{document}
\maketitle
\def\thefootnote{*}\footnotetext{Equal contribution}

\begin{abstract}
    Contact-rich manipulation, requiring robots to regulate not only motion but also how they yield to external forces, has emerged as the next frontier for Vision-Language-Action (VLA) models.
    However, existing VLAs output purely kinematic commands, degrading performance on real-world contact-rich tasks.
    In this paper, we introduce \textit{CompVLA}, a unified VLA framework that jointly predicts motion and a stiffness matrix from RGB and language inputs.
    Our approach augments the conventional architecture with a dedicated \textit{Compliance Expert}, which outputs time-varying stiffness and virtual displacement profiles executed via geometric impedance control.
    We demonstrate that CompVLA achieves the highest average success rate across diverse contact-rich tasks, outperforming both vanilla and compliance-aware VLA baselines, with ablations confirming each component is essential.
    See the project website at: \url{https://compvla.github.io}.

\end{abstract}


\section{Introduction}
\label{sec:introduction}

Vision-Language-Action (VLA) models~\cite{brohan2022rt, zitkovich2023rt, kim2024openvla, black2024pi_0, shukor2025smolvla} have become a major approach in robot learning. By mapping visual and language inputs directly to robot actions, and by training on diverse embodiments and tasks, VLAs extend the adaptability of foundation models to robotic control. However, most current VLAs use purely kinematic action representations. Their outputs specify target motion but typically leave the controller's interaction parameters, such as stiffness, unspecified. In contact-rich tasks such as polishing, peg-in-hole assembly, and whiteboard erasing, small tracking errors can cause the robot to lose contact or apply excessive force, which can damage the robot or the environment.

Compliance control addresses this issue by allowing the robot to yield during interaction~\cite{hogan1984impedance}. Rather than rigidly tracking a reference path, the robot absorbs position errors through controlled end-effector compliance. This reduces unsafe contact forces and improves robustness to execution errors. However, uniformly low stiffness is not always desirable. A stiffer controller tracks free-space motion more accurately. A task therefore needs variable compliance that stays stiff for precise free-space tracking and yields on contact, rather than a single fixed stiffness. Hand-tuning such profiles for every task does not scale, so prior work learns variable compliance from data~\cite{hou2025adaptive, kamijo2024learning, aburub2026learning}. Although these methods demonstrate the benefits of variable compliance in task-specific settings, its integration into VLAs remains an emerging research direction~\cite{zhang2026compliantvla}.

We aim to bring variable compliance to VLAs through two complementary capabilities: \emph{directional compliance} and \emph{force modulation}. Consider erasing a whiteboard. To follow a wiping path, the robot must stay stiff in the plane of motion while remaining compliant normal to the surface, since a single uniform stiffness is insufficient for complex contact tasks. This is directional compliance, remaining compliant along certain directions while staying stiff along others. Both these directions and their compliance vary across tasks and over the course of a task. Staying compliant, however, is not enough. Erasing also requires pressing the eraser into the board with a target force. This is force modulation. Compliance shapes the reaction to external forces but does not command pressing. When the target pose coincides with the actual contact pose, changing stiffness alone cannot generate a pressing force, since there is no spring deflection. Introducing a virtual displacement shifts the target into the surface, creating the deflection needed to apply a pressing force with the chosen stiffness.

\begin{figure}[tbp]
    \centering
    \includegraphics[width=1.0\linewidth]{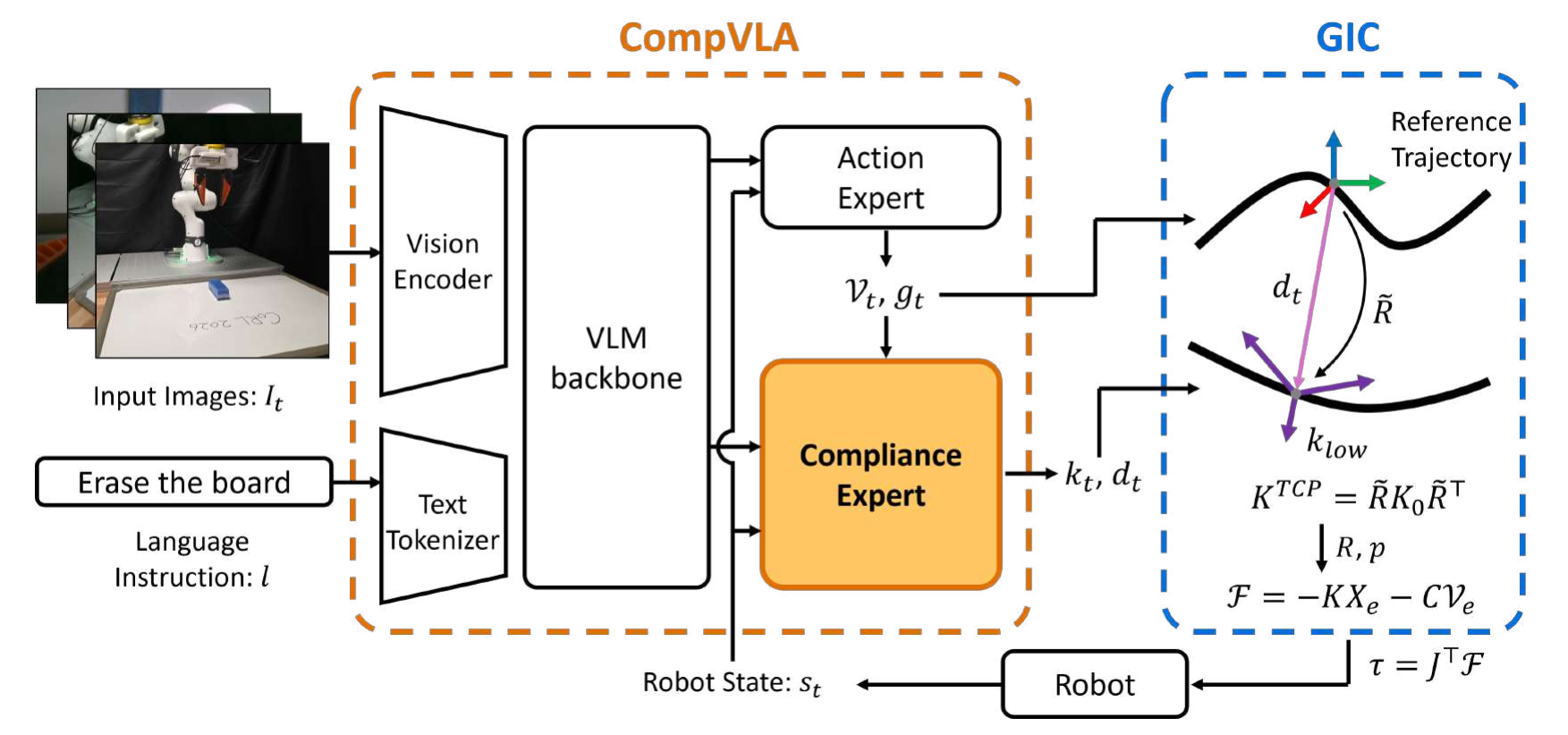}
    \caption{Overall framework of CompVLA and the geometric impedance controller. CompVLA outputs time-varying diagonal stiffness and virtual displacement vectors to realize directional compliance and force modulation.}
    \label{fig:CompVLA}
\end{figure}

We propose \textbf{CompVLA}, a VLA that learns variable compliance for contact-rich manipulation (Figure~\ref{fig:CompVLA}). It provides \emph{directional compliance} and \emph{force modulation} by adding a compliance expert to VLA architectures that pair a VLM with an action expert. Like the action expert, the compliance expert is conditioned on the VLM embedding, but it also takes the reference trajectory predicted by the action expert as input. It then predicts a time-varying diagonal stiffness and a virtual displacement. The diagonal stiffness entries set the compliance along each axis and the virtual displacement orients those axes, jointly providing directional compliance, while the same virtual displacement also generates the contact force the task requires, providing force modulation. Together with the action expert's desired trajectory, these outputs drive a geometric impedance controller adapted from~\cite{seo2023geometric,seo2024comparison} that produces coordinate-free compliant motion. This controller-aware design couples low-frequency motion generation with high-frequency interaction control through the predicted stiffness and virtual displacement.

We collect a teleoperated demonstration dataset with force/torque measurements and evaluate CompVLA on three challenging contact-rich tasks chosen to measure diverse capabilities, including directional compliance and force modulation. CompVLA outperforms existing baselines for contact-rich manipulation on average, including ones that use additional force~\cite{yu2026forcevla} or torque~\cite{zhang2025ta} inputs. It achieves these results without using external force/torque sensor measurements at inference. Contact-force measurements provide further evidence of CompVLA's force modulation capability. Ablations support the complementary roles of variable stiffness and virtual displacement in contact-rich manipulation. CompVLA also achieves performance comparable to $\pi_0$ on pick-and-place, a standard manipulation task. These results show that modeling compliant motion through variable stiffness and virtual displacement is an effective way to extend VLAs to physical interaction. We also release this dataset and the data conversion code to support future research on compliant manipulation.

\section{Related Work}
\label{sec:related_work}

\paragraph{Vision-Language-Action Models.}
Vision-Language-Action (VLA) models have advanced quickly by combining robot learning with large-scale vision-language priors. RT-1~\cite{brohan2022rt} learns tokenized robot actions, while RT-2~\cite{zitkovich2023rt} and OpenVLA~\cite{kim2024openvla} adapt pretrained VLMs to robot control using discretized action tokens. Large demonstration datasets such as Open X-Embodiment~\cite{o2024open} support training across diverse embodiments and tasks. Generative action heads offer an alternative to tokenized action representations by modeling continuous action distributions. Octo~\cite{mees2024octo} uses a diffusion head, while the $\pi_0$ family~\cite{black2024pi_0,black2025pi_} uses a flow-matching action expert. However, these models typically predict kinematic motion commands, with the controller treated as an external component. This limits the policy's ability to specify \emph{how} the robot physically interacts with the environment, a capability crucial for contact-rich manipulation.

\paragraph{VLA for Contact-Rich Manipulation.}
Recent work incorporates contact information into VLAs along two directions: \emph{input augmentation}, which adds contact signals while keeping pose-based action outputs, and \emph{output augmentation}, which expands the policy output itself. Input-augmentation methods introduce additional sensing modalities, such as a 6-axis wrench in ForceVLA~\cite{yu2026forcevla}, joint torque in TA-VLA~\cite{zhang2025ta} (which also predicts torque as an auxiliary output), and tactile images in VTLA~\cite{zhang2026vtla}. In all cases, these signals help the policy choose better target poses, while the policy does not explicitly specify objectives or parameters for interaction control beyond the motion reference. Output-augmentation methods are closer to our setting. ForceVLA2~\cite{li2026forcevla2} predicts a target wrench alongside the target pose via a Cross-Scale Mixture-of-Experts, supporting hybrid force-position control~\cite{raibert1981hybrid}. However, when high stiffness is used for accurate tracking along position-controlled directions, unexpected contact can generate large transient forces. For implementations that switch between position and force control, achieving smooth contact transitions also requires careful coordination of the two modes.

CompliantVLA-adaptor~\cite{zhang2026compliantvla} is the closest work to ours in that it handles contact through variable impedance control rather than mode switching. It attaches a separate VLM that selects stiffness and damping gains based on task execution phase and force/torque measurements. CompVLA instead predicts stiffness and virtual displacement with a Compliance Expert that shares the VLM backbone and learns these parameters from demonstrations. This design provides directional compliance and force modulation within a single model, without an additional VLM or external force/torque sensing at inference.

Our experiments include TA-VLA and ForceVLA as contact-aware VLA baselines. We exclude VTLA because our platform lacks tactile sensing. We also do not compare with ForceVLA2 or CompliantVLA-adaptor because their implementations are not publicly available.

\paragraph{Compliance in Imitation Learning.}
Classical compliance control~\cite{hogan1984impedance} has motivated a body of work on learning variable compliance~\cite{abu2020variable}, including stiffness learning for GIC~\cite{seo2023contact}. Prior methods estimate time-varying stiffness from human demonstrations using EMG~\cite{ajoudani2012tele}, demonstration variance~\cite{calinon2010learning,kronander2013learning}, or contact forces on the SPD manifold~\cite{abu2018force}. However, recovering both target pose and stiffness from a single demonstration is ill-posed because many pose--stiffness pairs yield the same trajectory and wrench. Comp-ACT~\cite{kamijo2024learning} and DIPCOM~\cite{aburub2026learning} avoid this inverse problem by collecting stiffness labels through teleoperator supervision, but such labels are costly to gather and depend on the teleoperator's intuition. Adaptive Compliance Policy (ACP)~\cite{hou2025adaptive} instead infers the compliance profile and target pose from position and wrench data via a heuristic. These methods are built on imitation-learning architectures such as diffusion policy~\cite{chi2025diffusion} and ACT~\cite{zhao2023learning}. CompVLA builds on this line of work by integrating a trajectory-conditioned Compliance Expert into a VLA. Adapting ACP's approach to compliance labeling, we derive stiffness and virtual displacement labels from force-feedback teleoperation demonstrations, and train CompVLA with a $\pi_0$ backbone.

\section{Preliminaries: Compliance Control}
\label{sec:preliminaries}

Compliance control specifies the end-effector's closed-loop behavior as a virtual mass--spring--damper,
\begin{equation}
\Lambda_{d}\, \ddot{\tilde{x}} + D_{d}\, \dot{\tilde{x}} + K_{d}\, \tilde{x} = F_{\mathrm{ext}},
\label{eq:msd-cartesian}
\end{equation}
where $\tilde{x} = x - x_{d}$ is the position error, $F_{\mathrm{ext}}$ the external force, and $\Lambda_{d}, D_{d}, K_{d} \succ 0$ the desired virtual inertia, damping, and stiffness. Intuitively, $K_{d}$ governs the trade-off between tracking $x_{d}$ and yielding to external forces, thereby shaping the force--displacement relationship. This model, however, is only an abstract Cartesian target, whereas a real manipulator moves on $\mathrm{SE}(3)$ and must be driven by joint torques. We now lift~\eqref{eq:msd-cartesian} to $\mathrm{SE}(3)$ and realize it at the actuator level.

\paragraph{Compliant behavior on $\mathrm{SE}(3)$.}
The Cartesian form~\eqref{eq:msd-cartesian} does not extend to $\mathrm{SE}(3)$: it treats $\tilde{x}$ as a vector, but $\mathrm{SE}(3)$ is not a vector space. We instead specify the behavior directly on $\mathrm{SE}(3)$, following the frame-invariant construction of~\citet{seo2024comparison}. Let $X = (R, p) \in \mathrm{SE}(3)$ be the end-effector pose, $X_{d} = (R_{d}, p_{d})$ its reference, and $\mathcal{V}^{b}, \mathcal{V}_d^{b} \in \mathbb{R}^{6}$ the body twists of $X$ and $X_d$. We express the reference twist in the body frame at $X$ as $\mathcal{V}_{d}^{\ast} = \mathrm{Ad}_{X^{-1}X_d}\mathcal{V}_{d}^{b}$, where $\mathrm{Ad}_{(\cdot)}$ denotes the adjoint map used here to transport the twist from the desired end-effector frame at $X_d$ to the current end-effector frame at $X$. The body-frame pose and twist errors are
\begin{equation}
X_{e} = \bigl(\log(X_{d}^{-1} X)\bigr)^{\vee},
\qquad
\mathcal{V}_{e} = \mathcal{V}^{b} - \mathcal{V}_{d}^{\ast},
\label{eq:se3-error}
\end{equation}
where the $(\cdot)^{\vee}$ map converts an element of $\mathfrak{se}(3)$ into a six-dimensional vector $[\omega; v] \in \mathbb{R}^{6}$ of its rotational and translational parts. The desired compliant behavior on $\mathrm{SE}(3)$ is
\begin{equation}
\Lambda_{d}\, \dot{\mathcal{V}}_{e} + D_{d}\, \mathcal{V}_{e} + K_{d}\, X_{e} = \mathcal{F}_{\mathrm{ext}},
\label{eq:msd-se3}
\end{equation}
with symmetric positive definite matrices $\Lambda_{d}, D_{d}, K_{d} \in \mathbb{R}^{6\times 6}$. The relative pose and body-frame twist errors in~\eqref{eq:se3-error} are invariant to a change of global reference frame, and so is the specification.

\paragraph{Realization via Impedance Control.}
We realize~\eqref{eq:msd-se3} through \textit{impedance control}~\cite{hogan1984impedance}, which commands joint torques $\tau$ that make the closed-loop dynamics match~\eqref{eq:msd-se3}.\footnote{An alternative is \emph{admittance control}, which uses position control: it integrates the measured wrench $\mathcal{F}_{\mathrm{ext}}$ through~\eqref{eq:msd-se3} into a position reference that an inner controller tracks. This suits position-controlled robots without a torque interface~\cite{pelletier1994implementation}.} The task-space dynamics are~\cite{lynch2017modern}
\begin{equation}
\Lambda(\theta)\, \dot{\mathcal{V}} + \eta(\theta, \mathcal{V}) = \mathcal{F} + \mathcal{F}_{\mathrm{ext}},
\qquad
\tau = J^{\top}(\theta)\, \mathcal{F},
\label{eq:os-dyn}
\end{equation}
where $\theta$ denotes the robot joint coordinates, $J(\theta)$ the Jacobian with $\mathcal{V} = J(\theta)\,\dot{\theta}$, $\Lambda(\theta)$ the task-space inertia, and $\eta(\theta, \mathcal{V}) = \mu(\theta, \mathcal{V})\,\mathcal{V} + \gamma(\theta)$ the centripetal and Coriolis terms plus gravity. For the following derivation, we express the dynamics in the current end-effector frame. Eliminating $\dot{\mathcal{V}}_{e}$ between~\eqref{eq:msd-se3} and~\eqref{eq:os-dyn} gives the task wrench
\begin{equation}
\mathcal{F} = \Lambda(\theta)\, \dot{\mathcal{V}}_{d}^{\ast}
+ \eta(\theta, \mathcal{V})
- \Lambda(\theta)\, \Lambda_{d}^{-1}\!\left(D_{d}\mathcal{V}_{e} + K_{d}X_e\right)
+ \bigl(\Lambda(\theta)\,\Lambda_{d}^{-1} - I\bigr)\mathcal{F}_{\mathrm{ext}}.
\label{eq:imp-law}
\end{equation}
This impedance formulation builds on \textit{geometric impedance control} (GIC)~\cite{seo2023geometric,seo2024comparison}, which provides the basis for our low-level controller.

Among the impedance parameters, we learn only the stiffness $K_{d}$, as in prior work~\cite{hou2025adaptive, kamijo2024learning, aburub2026learning}. This choice keeps the policy output compact while capturing the dominant term of the compliant behavior. In the quasistatic regime, where velocity and acceleration are negligible, \eqref{eq:msd-se3} reduces to $K_{d} X_{e} = \mathcal{F}_{\mathrm{ext}}$, so $K_{d}$ largely governs the force--displacement relationship. We set $\Lambda_{d} = \Lambda(\theta)$, which cancels the explicit external-wrench feedback term in~\eqref{eq:imp-law} and removes the need for force sensing in the control law. GIC compensates the Coriolis term along $\mathcal{V}_{d}^{\ast}$ instead of $\mathcal{V}$, which permits an energy-based proof of asymptotic stability for fixed gains and $\mathcal{F}_{\mathrm{ext}} = 0$~\cite{seo2024comparison}. In practice, CompVLA uses a simplified variant of this controller (Appendix~\ref{appendix:a2_control_detail}).

\section{Method}
\label{sec:method}

In this section, we describe how CompVLA learns to produce \emph{directional compliance} and \emph{force modulation}. We represent each motion's compliant behavior with a diagonal stiffness $k_t$ and a virtual displacement $d_t$, which together set how the robot yields along each axis and how it presses on the environment. We first describe how these targets are obtained from demonstrations. We then present the model that predicts them, followed by the controller that executes the predicted commands.

\subsection{Learning Compliance from Demonstration}
\label{subsec:learning_compliance}

We derive both capabilities from the interaction force and torque recorded during demonstration. We define $k_t = [k_t^R; k_t^p] \in \mathbb{R}^6$ and $d_t = [d_t^R; d_t^p] \in \mathbb{R}^6$, where the superscripts $R$ and $p$ denote the rotational and translational components, which we compute separately. We use the Tool Center Point (TCP) frame as the reference frame for execution. Directional compliance determines $k_t$ and the direction of $d_t$, and force modulation then sets the magnitude of $d_t$ given this stiffness. Figure~\ref{fig:compliance_analysis} in Appendix~\ref{appendix:c2_visualization} shows both behaviors in CompVLA's predictions during execution, with directional compliance during Pivot in (a) and force modulation during Cup Dispense in (b).

\paragraph{Directional Compliance.}
The robot should yield along the contact direction, where reaction forces concentrate, while remaining stiff in the orthogonal directions to preserve tracking. We estimate this direction from the measured interaction force or torque and set the direction of $d_t$ accordingly. For each translational and rotational space, we define a compliance frame whose first axis is aligned with this direction. The remaining rotational freedom of each frame is resolved by choosing the orientation closest to the TCP frame along the shortest geodesic path. In this frame, we assign a low stiffness $k_{\text{low}}$ to the contact axis and a high stiffness $k_{\text{max}}$ to the two orthogonal axes, yielding $K_0 = \text{diag}(k_{\text{low}}, k_{\text{max}}, k_{\text{max}})$. We adapt the force-dependent stiffness rule of Adaptive Compliance Policy~\cite{hou2025adaptive} to the translational and rotational spaces, setting $k_{\text{low}}$ from the corresponding force or torque magnitude to obtain the stiffness target $k_t$. Together, $k_t$ and $d_t$ define the direction and magnitude of the anisotropic compliance used by the controller. The stiffness-labeling rule is detailed in Appendix~\ref{appendix:b2_learning_k}.

\paragraph{Force Modulation.}
To modulate the pressing force while retaining the chosen stiffness profile, we use $d_t$ to shift the equilibrium pose relative to the reference trajectory. This shift generates an elastic force or torque through the spring term of the impedance controller. Recovering $(k_t, d_t)$ from a target force or torque is ill-posed, since many pairs can produce the same value. With the contact direction and $k_{\text{low}}$ fixed as above, we set the displacement magnitude to the demonstrated force or torque magnitude divided by $k_{\text{low}}$, treating translation and rotation separately. Thus, $d_t$ provides both the direction of the virtual shift and the displacement magnitude for force generation, while $k_t$ sets the corresponding stiffness.

\subsection{CompVLA}

We now describe the policy architecture that predicts the compliance-aware action representation defined above. CompVLA extends a standard VLA policy with a compliance branch that predicts the stiffness vector $k_t$ and virtual displacement vector $d_t$ in addition to motion commands. We first define the policy inputs and motion tokens, then describe the model architecture and how its outputs generate compliant motion via GIC.

\paragraph{Problem Formulation.}
At each timestep $t$, CompVLA receives a language instruction $l$ and an observation $O_t = \{I_t, s_t\}$, where $I_t$ denotes camera images and $s_t$ denotes the robot state. The state includes the gripper state and the current Tool Center Point (TCP) pose $X_t = (R_t, p_t) \in \mathrm{SE}(3)$. External force/torque measurements are used to construct the compliance labels and are not required at inference. From these inputs, the policy $\pi(M_t \mid O_t, l)$ predicts a motion chunk over a horizon of $T$ steps:
\begin{equation}\label{eq:motion_chunk}
    M_t = (m_t, m_{t+1}, \dots, m_{t+T-1}).
\end{equation}
Each motion token $m_t \in M_t$ is parameterized as
\begin{equation}\label{eq:motion_token}
    m_t = (\mathcal{V}_t, g_t, k_t, d_t),
\end{equation}
where $\mathcal{V}_t \in \mathbb{R}^6$ is the reference TCP twist and $g_t \in \{0, 1\}$ is the gripper command. The stiffness vector $k_t \in \mathbb{R}^6$ specifies the diagonal stiffness, and the virtual displacement $d_t \in \mathbb{R}^6$ specifies the direction and amount of virtual target shift. Together, they parameterize the controller's anisotropic compliance and force modulation.

\paragraph{Model Architecture.}
As illustrated in Figure~\ref{fig:CompVLA}, CompVLA builds upon a core Vision-Language Model (VLM) coupled with a dual-expert structure consisting of a \textit{Action Expert} and a \textit{Compliance Expert}. Both experts attend to the VLM's key-value representations using a blockwise causal attention mask. At inference, given the visual inputs, language instruction, and robot state, the Action Expert first predicts the reference TCP twist $\mathcal{V}_t$ and the binary gripper command $g_t$. This predicted twist is integrated to generate a reference trajectory, which then serves as an explicit conditioning signal for the Compliance Expert. Processing this trajectory alongside the VLM context, the Compliance Expert outputs the diagonal stiffness $k_t$ and the virtual displacement $d_t$. Finally, the generated reference trajectory, diagonal stiffness, and virtual displacement are transmitted to the downstream controller for physical execution. During training, the Compliance Expert is conditioned on the ground-truth TCP trajectory. We jointly train the VLM backbone and both experts using flow matching, as detailed in Appendix~\ref{appendix:a1_network_detail}.

\paragraph{Compliant Motion Generation.}
Upon receiving the predicted motion chunk, the robot executes it with a high-frequency GIC. Impedance controllers in robot-learning setups often use scalar or diagonal stiffness gains in a predefined control frame, without explicitly adapting the compliance axes to the contact direction. In contrast, CompVLA uses the predicted virtual displacement $d_t$ both to shift the reference trajectory and to adapt the orientation of the compliance frame.

Specifically, the reference trajectory is shifted by $d_t$ to form a virtual target pose $X^{v}_{t} = (R^{v}_{t}, p^{v}_{t}) \in \mathrm{SE}(3)$. Tracking this virtual target induces force modulation through the spring term of the impedance controller. At the same time, the direction of $d_t$ determines the primary compliance axis, enabling directional compliance as described above. Let $\tilde{R}_t^R, \tilde{R}_t^p \in \mathrm{SO}(3)$ denote the orientations of the rotational and translational compliance frames with respect to the TCP frame. We use these rotations to map the predicted diagonal stiffness values into the TCP frame and construct the task-space stiffness matrix $K_t$. The resulting stiffness matrix and virtual target pose parameterize the spring--damper controller detailed in Appendix~\ref{appendix:a2_control_detail}.

\section{Experiments}
\label{sec:experiments}
\subsection{Experiment Setup}

\begin{wrapfigure}{r}{0.38\textwidth}
    \centering
    \vspace{-13pt}
    \includegraphics[width=1\linewidth]{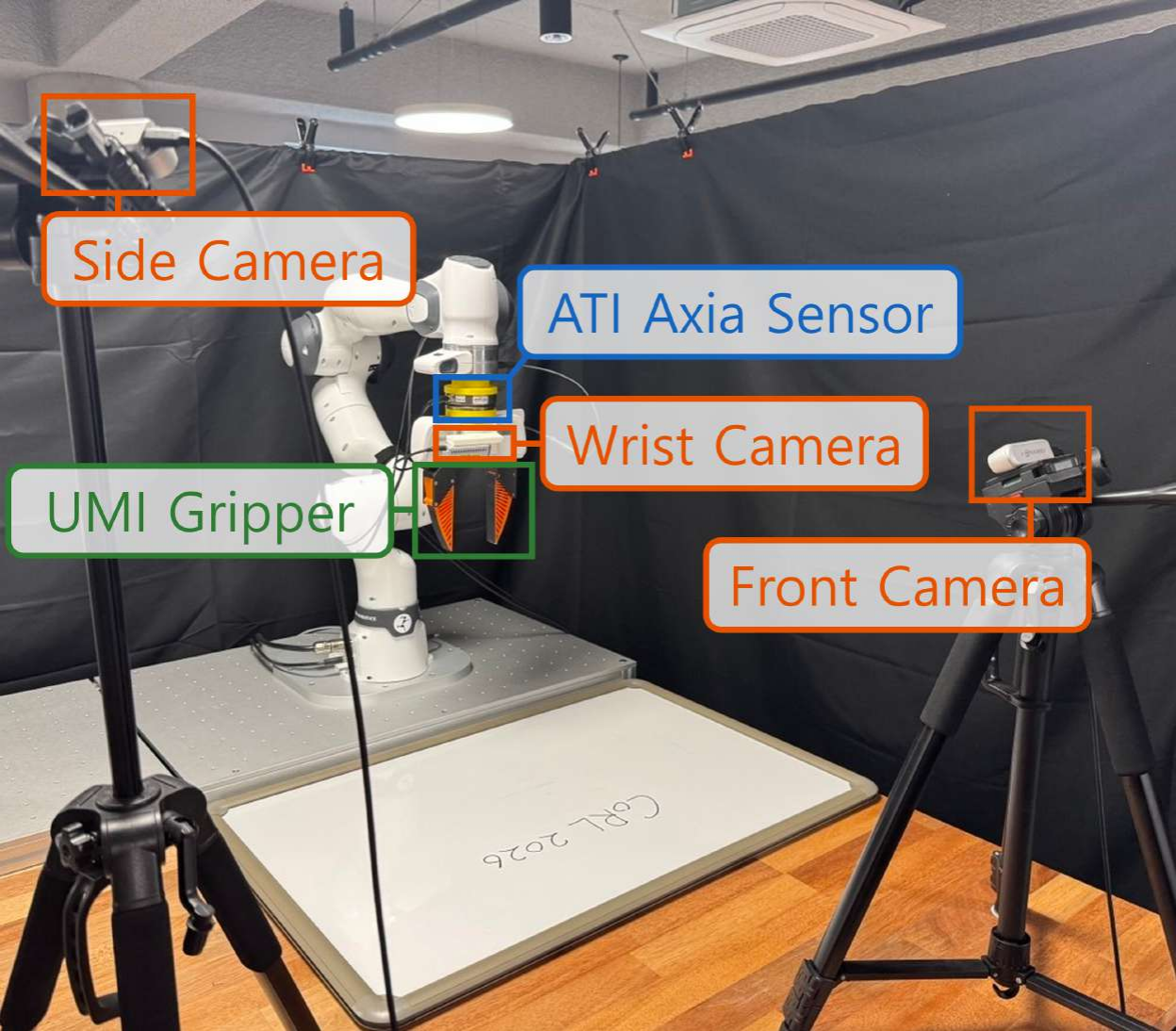}
    \vspace{-13pt}
    \caption{Hardware setup used for our experiments.}
    \vspace{-12pt}
    \label{fig:hardware_setup}
\end{wrapfigure}

\paragraph{Hardware Setup.}
Our hardware platform consists of a force-aware leader-follower teleoperation system built around two 7-DoF Franka Research 3 (FR3) arms, shown in Figure~\ref{fig:hardware_setup}. The follower arm, used for both data collection and evaluation, is controlled via a task-space compliance controller. It is equipped with three Intel RealSense D435i cameras mounted at the front, side, and wrist, providing complementary viewpoints for policy learning. We replace the default FR3 fingers with the UMI fingertip design~\cite{chi2024universal} to improve grasping performance. An ATI Axia force/torque sensor is mounted at the wrist; while its readings are recorded during data collection, they are not used by CompVLA's policy or controller at evaluation time. From each demonstration, we collect camera images from all three cameras, joint positions, joint torques, and force/torque sensor readings. We release the raw contact-rich demonstrations and the data conversion code to support future research on compliant manipulation. Dataset processing is detailed in Appendix~\ref{appendix:b3_dataset_conversion}.

\begin{figure}[tbp]
    \centering
    \includegraphics[width=1.0\linewidth]{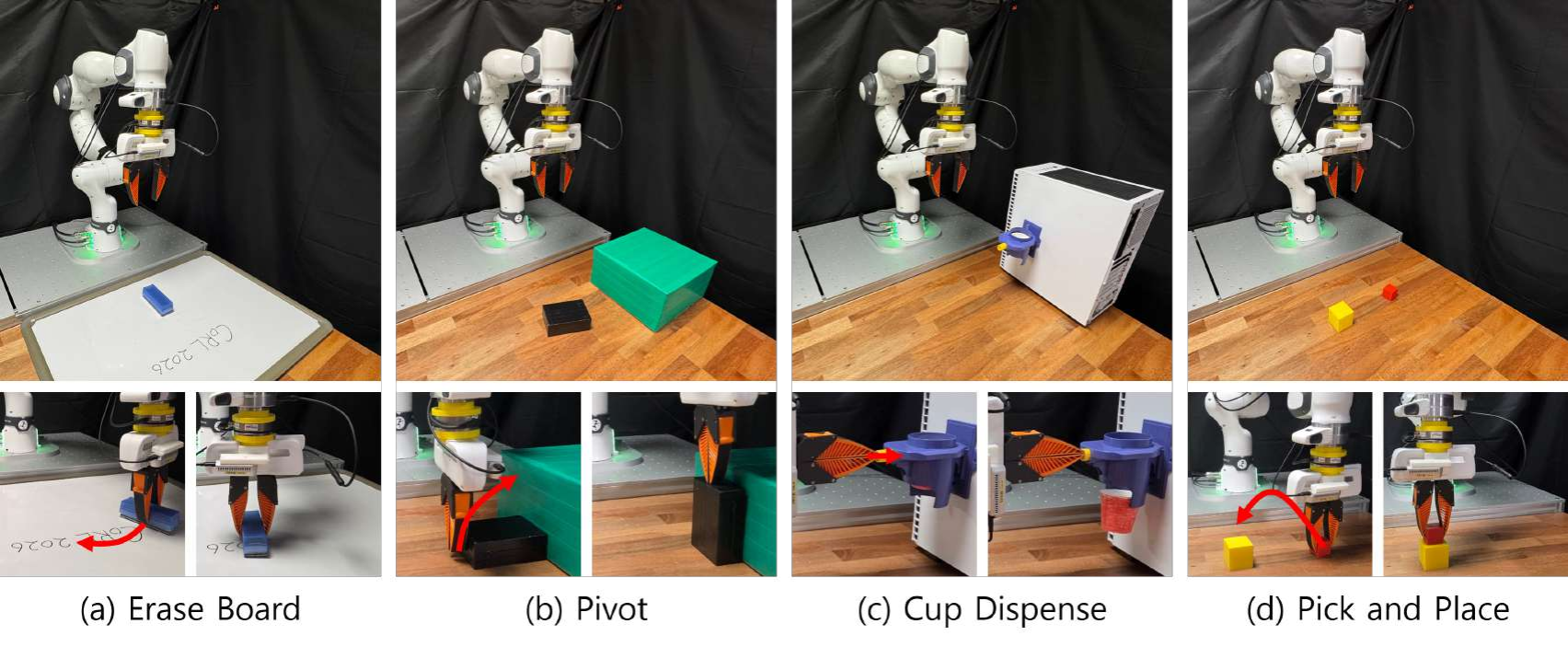}
    \caption{Overview of tasks used in the experiments. Tasks (a)--(c) are three cases where compliance is essential. Task (d), Pick and Place, verifies that CompVLA is also capable of standard manipulation.}
    \label{fig:tasks}
\end{figure}

\paragraph{Tasks.}
To evaluate the compliance control capability of the VLA model, we conduct experiments on three contact-rich tasks, shown in Figure~\ref{fig:tasks}: Erase Whiteboard, Pivot, and Cup Dispense. These tasks span the regimes in which compliance is essential: surface tracking under pose uncertainty (Erase), continuously varying compliance direction (Pivot), and force-regulated pressing to a precise depth (Cup Dispense)~\cite{yu2026forcevla}. Success is binary for all tasks except Erase, for which we report the fraction of letters successfully erased from the board. We additionally include a Pick and Place task (Figure~\ref{fig:tasks}(d)) to evaluate CompVLA on standard manipulation. For each task, we collect 100 demonstrations for training and evaluate over 20 episodes with randomized environment initialization. For each VLA method, we conduct three independent evaluation runs using the same trained model and report the mean and standard deviation across runs. An episode terminates if it exceeds a task-specific time limit or triggers the FR3's built-in safety stop due to excessive contact force.

\paragraph{Baselines.}
We compare CompVLA against three VLA baselines: $\pi_0$~\cite{black2024pi_0}, a typical VLA model, and two of its variants designed for contact-rich manipulation, ForceVLA~\cite{yu2026forcevla} and TA-VLA~\cite{zhang2025ta}. We modified ForceVLA to accept three camera views instead of its original two, matching our three-camera setup. For TA-VLA, we adopt the \textit{DePost-1} architecture with the unified training objective that uses future torques as an auxiliary training signal alongside action prediction, following the configuration evaluated in the original paper. For fair comparison, we also leverage $\pi_0$ as a backbone of CompVLA. We jointly fine-tune the VLM backbone and policy experts for CompVLA and the VLA baselines. We additionally evaluate Diffusion Policy (DP)~\cite{chi2025diffusion} and ACT~\cite{zhao2023learning} as non-VLA imitation-learning baselines. All baselines track their predicted motion using the same GIC, with stiffness fixed at the maximum values used by CompVLA.

\begin{table}[tbp]
    \centering
    \caption{
    Success rates (\%). VLA results are mean $\pm$ standard deviation over three evaluation runs; DP and ACT use one run. Each run contains 20 episodes per task. Erase uses character-based success; other tasks use episode success. Average covers the three contact-rich tasks.
    }
    \label{table:main_results}
    \begin{tabular}{@{}l|cccc|c@{}}
        \toprule
        \textbf{Method} & \multicolumn{4}{c|}{\textbf{Contact-rich}} & \textbf{Standard} \\
        \cmidrule(lr){2-5} \cmidrule(lr){6-6}
        \textbf{Task} & Erase & Pivot & Cup Dispense & Average & PnP \\
        \midrule
        $\pi_0$~\cite{black2024pi_0} & \ms{17.8}{7.1} & \ms{31.7}{10.4} &
        \ms{45.0}{0.0} &
        \ms{31.5}{5.2} &
        \ms{80.0}{10.0} \\
        ForceVLA~\cite{yu2026forcevla} & \ms{68.1}{4.1} & \ms{45.0}{15.0} & \msb{50.0}{8.7} & \ms{54.4}{4.3} & \ms{45.0}{15.0} \\
        TA-VLA~\cite{zhang2025ta} & \ms{52.8}{9.7} & \ms{26.7}{10.4} & \ms{0.0}{0.0} & \ms{26.5}{2.0} & \msb{83.3}{12.6} \\
        CompVLA (ours) & \msb{92.9}{3.7} & \msb{73.3}{10.4} & \ms{35.0}{10.0} & \msb{67.1}{7.2} & \ms{76.7}{5.8} \\
        \midrule
        DP~\cite{chi2025diffusion} & 0.0 & 0.0 & 10.0 & 3.3 & 50.0 \\
        ACT~\cite{zhao2023learning} & 0.0 & 15.0 & 30.0 & 15.0 & 0.0 \\
        \bottomrule
    \end{tabular}
\end{table}

\subsection{Results}

Table~\ref{table:main_results} reports the success rates of CompVLA and the baselines. On the contact-rich tasks, CompVLA achieves an average success rate of \textbf{67.1\%}, outperforming the strongest baseline by 12.7 percentage points. Although ForceVLA and TA-VLA try to incorporate force/torque information to enhance performance in contact-rich tasks, their position-only action outputs do not directly command compliance or contact forces during execution. The additional non-VLA baselines, DP and ACT, achieve average contact-rich success rates of 3.3\% and 15.0\%, respectively, in their single-run evaluations. Despite this overall advantage, CompVLA's success on Cup Dispense remains limited. Additional stage-wise evaluations indicate that accurately reaching the small button is the main bottleneck, while CompVLA achieves higher pushing success than $\pi_0$ (Appendix~\ref{appendix:c1_contact_analysis}).

\begin{figure}[tbp]
    \centering
    \includegraphics[width=1.0\linewidth]{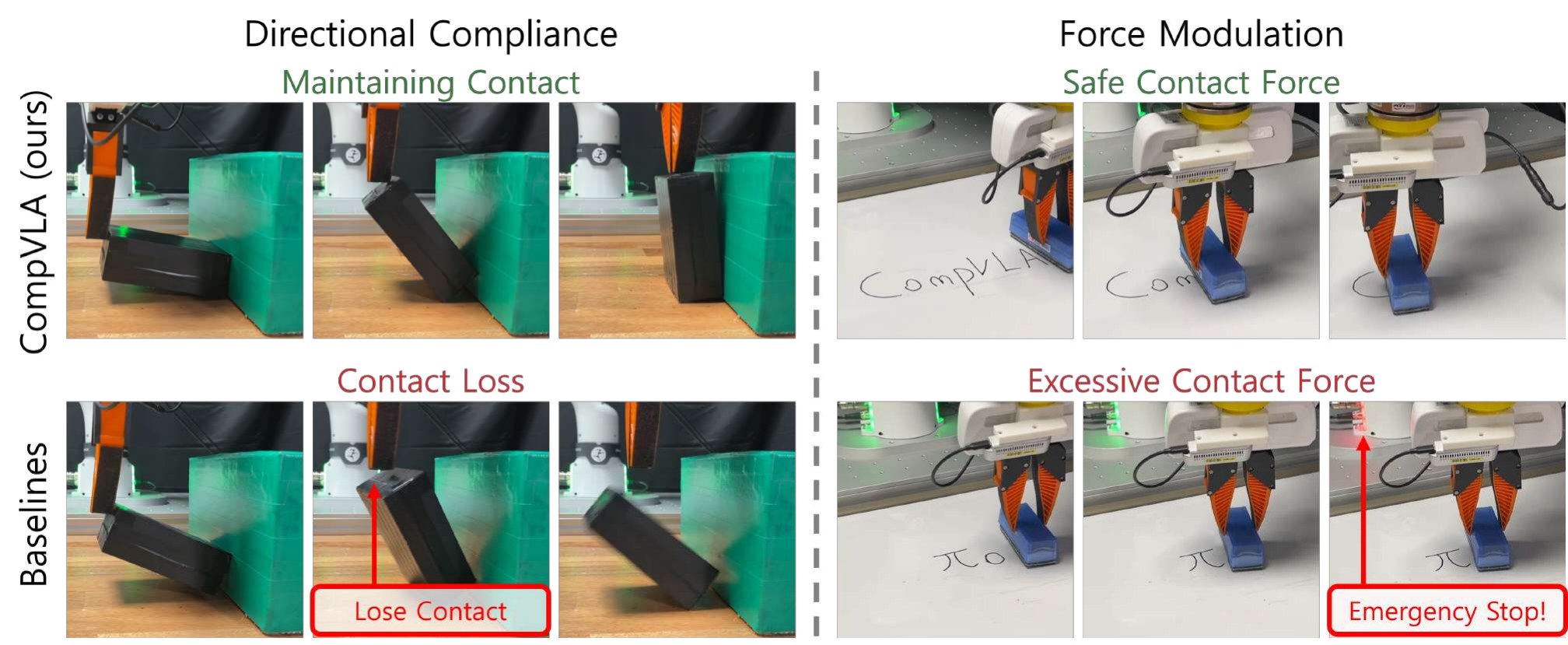}
    \caption{Qualitative comparison of CompVLA and position-only baseline rollouts.
    CompVLA (top) leverages directional compliance to maintain contact as the contact direction changes (left), and force modulation to apply controlled forces (right).
    In contrast, position-only baselines (bottom) exhibit two characteristic failures: losing contact with the manipulated object (left), and pressing too aggressively into the surface, triggering the FR3's safety stop (right).}
    \label{fig:failure_case}
\end{figure}

As shown in Figure~\ref{fig:failure_case}, baseline rollouts typically fail in one of two modes. In the first case, illustrated here by the Pivot task, the end-effector drifts away from the black box and fails to tip it. In the second case, illustrated here by the Erase Whiteboard task, the robot presses too aggressively into the environment, triggering the FR3's built-in safety stop due to excessive contact force. In contrast, CompVLA's explicit stiffness prediction lets the policy modulate compliance on the fly, maintaining stable contact while keeping interaction forces within safe limits.

Additionally, the rightmost column of Table~\ref{table:main_results} reports performance on Pick and Place, a standard manipulation task that does not require compliance. CompVLA achieves performance comparable to $\pi_0$ (76.7\% vs. 80.0\%), suggesting that it remains effective even when compliance is unnecessary. In our ForceVLA implementation, we observed oscillations near discontinuities in the predicted Euler angles, particularly on Pick and Place, which contributed to failures.

\vspace{20pt}

\subsection{Ablation Study}

In this section, we conduct ablation studies to validate (i) the contribution of each component in the Compliance Expert and (ii) the benefit of variable compliance over a fixed stiffness matrix.

\begin{wraptable}{r}{0.4\linewidth}
    \centering
    \caption{Ablation of the Compliance Expert components on the Erase Whiteboard task.}
    \label{tab:remove_comp}
    {
    \footnotesize
    \begin{tabular}{l|c}
        \toprule
        Method & Erase Whiteboard \\
        \midrule
        Comp w/o virtual & 7.0\% \\
        Comp w/o $K$\&virtual & 15.3\% \\
        CompVLA (ours) & \textbf{96.0\%} \\
        \bottomrule
    \end{tabular}
    }
\end{wraptable}

\paragraph{Removing Components of the Compliance Module.}
We evaluate two ablations of CompVLA on the Erase Whiteboard task with 10 episodes. \textit{Comp w/o virtual} removes only the virtual target shift during evaluation, retaining the predicted stiffness and compliance-frame orientation, and \textit{Comp w/o $K$\&virtual} removes the Compliance Expert entirely, resulting in a position-only policy. As reported in Table~\ref{tab:remove_comp}, removing these components drops the success rate from 96.0\% to 7.0\% and 15.3\%, respectively. \textit{Comp w/o virtual} performs even worse than the position-only \textit{Comp w/o $K$\&virtual} because the stiffness is jointly trained to complement the virtual displacement. Using this stiffness without its paired displacement introduces a position bias and leads to insufficient contact force. These results confirm that the two predictions must be used together for reliable contact interaction.

\paragraph{Fixed Stiffness Matrix $K$.}
\begin{wraptable}{r}{0.4\linewidth}
    \centering
    \caption{Ablation with a fixed stiffness matrix on the Pivot task.}
    \label{tab:fixk}
    {
    \footnotesize
    \begin{tabular}{l|c}
        \toprule
        Method & Pivot \\
        \midrule
        Fixed $K$ & 40\% \\
        CompVLA (ours) & \textbf{70\%} \\
        \bottomrule
    \end{tabular}
    }
\end{wraptable}

Variable compliance is a core design choice in CompVLA. We fix the diagonal stiffness entries to $30\,\mathrm{N\,m/rad}$ for rotation and $500\,\mathrm{N/m}$ for translation, retaining the predicted virtual displacement $d_t$, and evaluate 10 Pivot episodes. As shown in Table~\ref{tab:fixk}, fixing the stiffness matrix reduces the success rate from 70\% to 40\%, highlighting the importance of adaptive stiffness for handling tasks with changing compliance requirements.

\section{Conclusion}
\label{sec:conclusion}

We present \textit{CompVLA}, a vision-language-action policy that enables physical interaction by predicting reference motion together with diagonal stiffness and virtual displacement profiles. When integrated with GIC, CompVLA connects low-frequency multimodal policy outputs with a high-frequency compliant control loop. Through a dual-expert architecture, the policy predicts reference TCP motion, stiffness, and virtual displacement commands, which are used to form virtual target poses and task-space stiffness matrices for directional compliance and force modulation. Robot experiments across diverse tasks such as whiteboard erasing, pivoting, and cup dispensing demonstrate that CompVLA improves average performance over kinematics-only and force-conditioned baselines, while preserving standard manipulation performance in pick-and-place routines. Ultimately, these results show that incorporating directional compliance and force modulation into VLA action representations is an effective way to extend foundation models to contact-rich environments.

\section{Limitations and Future Work}
\label{sec:limitations}

One limitation of CompVLA is that its compliance behavior depends on the quality of the predicted reference motion. The compliance expert can regulate contact once the end-effector reaches the interaction region, but it cannot fully compensate for large trajectory errors. This limitation is most visible in the cup dispensing task, where the robot must reach a small button before force modulation can be effective. In contrast, erasing and pivoting have broader contact regions and are more tolerant to moderate pose errors. This suggests that CompVLA would benefit from stronger reference motion generation, especially for tasks with small contact targets or narrow affordance regions.

Another limitation comes from how the compliance targets are defined. We derive them from force/torque measurements, which works well after contact is established but provides limited guidance for defining the desired compliance before contact. The current approach also focuses on a dominant compliance direction, which may be insufficient for tasks that require compliance across multiple directions or strongly coupled rotational and translational behavior. Future work should reduce the reliance on force-derived labels, explore richer stiffness representations, and scale the data to more diverse contact conditions.
\clearpage

\acknowledgments{
J. Kim, J. Ha, C.-S. Park, M. Song, H. Jeong, H. Hwang, and F. C. Park were supported 
in part by IITP-MSIT under Grant RS-2024-00436680 (Collaborative Research Projects with Microsoft Research) through Global Research Support Program in the Digital Field program, 
in part by IITP-MSIT under Grant 2022-220480 and Grant RS-2022-II220480 (Training and Inference Methods for Goal Oriented AI Agents), 
in part by KIAT under Grant P0020536 (HRD Program for Industrial Innovation), 
in part by MOTIR under Grant RS-2025-25462891, 
in part by MOTIR under Grant RS-2025-25460896, 
in part by SNU-IPAI,
in part by SNU-AIIS,
in part by SNU-IAMD,
in part by SNU BK21+ Program in Mechanical Engineering, 
in part by SNU Institute for Engineering Research,
in part by Hyundai Motor Company and Kia,
and in part by Microsoft Research Asia.
}
\bibliography{references}  

@article{brohan2022rt,
  title={Rt-1: Robotics transformer for real-world control at scale},
  author={Brohan, Anthony and Brown, Noah and Carbajal, Justice and Chebotar, Yevgen and Dabis, Joseph and Finn, Chelsea and Gopalakrishnan, Keerthana and Hausman, Karol and Herzog, Alex and Hsu, Jasmine and others},
  journal={arXiv preprint arXiv:2212.06817},
  year={2022}
}

@inproceedings{zitkovich2023rt,
  title={Rt-2: Vision-language-action models transfer web knowledge to robotic control},
  author={Zitkovich, Brianna and Yu, Tianhe and Xu, Sichun and Xu, Peng and Xiao, Ted and Xia, Fei and Wu, Jialin and Wohlhart, Paul and Welker, Stefan and Wahid, Ayzaan and others},
  booktitle={Conference on Robot Learning},
  pages={2165--2183},
  year={2023},
  organization={PMLR}
}

@article{kim2024openvla,
  title={Openvla: An open-source vision-language-action model},
  author={Kim, Moo Jin and Pertsch, Karl and Karamcheti, Siddharth and Xiao, Ted and Balakrishna, Ashwin and Nair, Suraj and Rafailov, Rafael and Foster, Ethan and Lam, Grace and Sanketi, Pannag and others},
  journal={arXiv preprint arXiv:2406.09246},
  year={2024}
}

@article{black2024pi_0,
  title={$\pi_0 $: A Vision-Language-Action Flow Model for General Robot Control},
  author={Black, Kevin and Brown, Noah and Driess, Danny and Esmail, Adnan and Equi, Michael and Finn, Chelsea and Fusai, Niccolo and Groom, Lachy and Hausman, Karol and Ichter, Brian and others},
  journal={arXiv preprint arXiv:2410.24164},
  year={2024}
}

@article{shukor2025smolvla,
  title={Smolvla: A vision-language-action model for affordable and efficient robotics},
  author={Shukor, Mustafa and Aubakirova, Dana and Capuano, Francesco and Kooijmans, Pepijn and Palma, Steven and Zouitine, Adil and Aractingi, Michel and Pascal, Caroline and Russi, Martino and Marafioti, Andres and others},
  journal={arXiv preprint arXiv:2506.01844},
  year={2025}
}

@inproceedings{hogan1984impedance,
  title={Impedance control: An approach to manipulation},
  author={Hogan, Neville},
  booktitle={1984 American control conference},
  pages={304--313},
  year={1984},
  organization={IEEE}
}

@inproceedings{hou2025adaptive,
  title={Adaptive compliance policy: Learning approximate compliance for diffusion guided control},
  author={Hou, Yifan and Liu, Zeyi and Chi, Cheng and Cousineau, Eric and Kuppuswamy, Naveen and Feng, Siyuan and Burchfiel, Benjamin and Song, Shuran},
  booktitle={2025 IEEE International Conference on Robotics and Automation (ICRA)},
  pages={4829--4836},
  year={2025},
  organization={IEEE}
}

@inproceedings{kamijo2024learning,
  title={Learning variable compliance control from a few demonstrations for bimanual robot with haptic feedback teleoperation system},
  author={Kamijo, Tatsuya and Beltran-Hernandez, Cristian C and Hamaya, Masashi},
  booktitle={2024 IEEE/RSJ International Conference on Intelligent Robots and Systems (IROS)},
  pages={12663--12670},
  year={2024},
  organization={IEEE}
}

@inproceedings{aburub2026learning,
  title={Learning diffusion policies from demonstrations for compliant contact-rich manipulation},
  author={Aburub, Malek and Beltran-Hernandez, Cristian C and Kamijo, Tatsuya and Hamaya, Masashi},
  booktitle={2026 IEEE/SICE International Symposium on System Integration (SII)},
  pages={28--34},
  year={2026},
  organization={IEEE}
}

@article{zhang2026compliantvla,
  title={Compliantvla-adaptor: Vlm-guided variable impedance action for safe contact-rich manipulation},
  author={Zhang, Heng and Huang, Wei-Hsing and Tong, Qiyi and Solak, Gokhan and Liu, Puze and Zhang, Kaidi and Liu, Sheng and Peters, Jan and She, Yu and Ajoudani, Arash},
  journal={arXiv preprint arXiv:2601.15541},
  year={2026}
}

@article{seo2023geometric,
  title={Geometric impedance control on SE (3) for robotic manipulators},
  author={Seo, Joohwan and Prakash, Nikhil Potu Surya and Rose, Alexander and Choi, Jongeun and Horowitz, Roberto},
  journal={IFAC-PapersOnLine},
  volume={56},
  number={2},
  pages={276--283},
  year={2023},
  publisher={Elsevier}
}

@inproceedings{seo2024comparison,
  title={A comparison between lie group-and lie algebra-based potential functions for geometric impedance control},
  author={Seo, Joohwan and Prakash, Nikhil Potu Surya and Choi, Jongeun and Horowitz, Roberto},
  booktitle={2024 American Control Conference (ACC)},
  pages={1335--1342},
  year={2024},
  organization={IEEE}
}

@article{yu2026forcevla,
  title={Forcevla: Enhancing vla models with a force-aware moe for contact-rich manipulation},
  author={Yu, Jiawen and Liu, Hairuo and Yu, Qiaojun and Ren, Jieji and Hao, Ce and Ding, Haitong and Huang, Guangyu and Huang, Guofan and Song, Yan and Cai, Panpan and others},
  journal={Advances in Neural Information Processing Systems},
  volume={38},
  pages={93409--93439},
  year={2026}
}

@article{zhang2025ta,
  title={Ta-vla: Elucidating the design space of torque-aware vision-language-action models},
  author={Zhang, Zongzheng and Xu, Haobo and Yang, Zhuo and Yue, Chenghao and Lin, Zehao and Gao, Huan-ang and Wang, Ziwei and Zhao, Hao},
  journal={arXiv preprint arXiv:2509.07962},
  year={2025}
}

@inproceedings{o2024open,
  title={Open x-embodiment: Robotic learning datasets and rt-x models: Open x-embodiment collaboration 0},
  author={O’Neill, Abby and Rehman, Abdul and Maddukuri, Abhiram and Gupta, Abhishek and Padalkar, Abhishek and Lee, Abraham and Pooley, Acorn and Gupta, Agrim and Mandlekar, Ajay and Jain, Ajinkya and others},
  booktitle={2024 IEEE International Conference on Robotics and Automation (ICRA)},
  pages={6892--6903},
  year={2024},
  organization={IEEE}
}

@inproceedings{mees2024octo,
  title={Octo: An open-source generalist robot policy},
  author={Mees, Oier and Ghosh, Dibya and Pertsch, Karl and Black, Kevin and Walke, Homer Rich and Dasari, Sudeep and Hejna, Joey and Kreiman, Tobias and Xu, Charles and Luo, Jianlan and others},
  booktitle={First Workshop on Vision-Language Models for Navigation and Manipulation at ICRA 2024},
  year={2024}
}

@inproceedings{black2025pi_,
  title={$\pi_{0.5}$: a Vision-Language-Action Model with Open-World Generalization},
  author={Black, Kevin and Brown, Noah and Darpinian, James and Dhabalia, Karan and Driess, Danny and Esmail, Adnan and Equi, Michael Robert and Finn, Chelsea and Fusai, Niccolo and Galliker, Manuel Y and others},
  booktitle={9th Annual Conference on Robot Learning},
  year={2025}
}

@article{zhang2026vtla,
  title={Vtla: Vision-tactile-language-action model with preference learning for insertion manipulation},
  author={Zhang, Chaofan and Hao, Peng and Cao, Xiaoge and Hao, Xiaoshuai and Cui, Shaowei and Wang, Shuo},
  journal={Biomimetic Intelligence and Robotics},
  pages={100333},
  year={2026},
  publisher={Elsevier}
}

@article{li2026forcevla2,
  title={ForceVLA2: Unleashing Hybrid Force-Position Control with Force Awareness for Contact-Rich Manipulation},
  author={Li, Yang and Jiang, Hongru and Xia, Junjie and Zhang, Hongquan and Du, Jinda and Zhou, Yunsong and Zeng, Jia and Hao, Ce and Ren, Jieji and Yu, Qiaojun and others},
  journal={arXiv preprint arXiv:2603.15169},
  year={2026}
}

@article{raibert1981hybrid,
  title={Hybrid position/force control of manipulators},
  author={Raibert, Marc H and Craig, John J},
  year={1981}
}

@article{abu2020variable,
  title={Variable impedance control and learning—a review},
  author={Abu-Dakka, Fares J and Saveriano, Matteo},
  journal={Frontiers in Robotics and AI},
  volume={7},
  pages={590681},
  year={2020},
  publisher={Frontiers Media SA}
}

@article{seo2023contact,
  title={Contact-rich se (3)-equivariant robot manipulation task learning via geometric impedance control},
  author={Seo, Joohwan and Prakash, Nikhil PS and Zhang, Xiang and Wang, Changhao and Choi, Jongeun and Tomizuka, Masayoshi and Horowitz, Roberto},
  journal={IEEE Robotics and Automation Letters},
  volume={9},
  number={2},
  pages={1508--1515},
  year={2023},
  publisher={IEEE}
}

@article{ajoudani2012tele,
  title={Tele-impedance: Teleoperation with impedance regulation using a body--machine interface},
  author={Ajoudani, Arash and Tsagarakis, Nikos and Bicchi, Antonio},
  journal={The International Journal of Robotics Research},
  volume={31},
  number={13},
  pages={1642--1656},
  year={2012},
  publisher={SAGE Publications Sage UK: London, England}
}

@inproceedings{calinon2010learning,
  title={Learning-based control strategy for safe human-robot interaction exploiting task and robot redundancies},
  author={Calinon, Sylvain and Sardellitti, Irene and Caldwell, Darwin G},
  booktitle={2010 IEEE/RSJ International Conference on Intelligent Robots and Systems},
  pages={249--254},
  year={2010},
  organization={IEEE}
}

@article{kronander2013learning,
  title={Learning compliant manipulation through kinesthetic and tactile human-robot interaction},
  author={Kronander, Klas and Billard, Aude},
  journal={IEEE transactions on haptics},
  volume={7},
  number={3},
  pages={367--380},
  year={2013},
  publisher={IEEE}
}

@article{abu2018force,
  title={Force-based variable impedance learning for robotic manipulation},
  author={Abu-Dakka, Fares J and Rozo, Leonel and Caldwell, Darwin G},
  journal={Robotics and Autonomous Systems},
  volume={109},
  pages={156--167},
  year={2018},
  publisher={Elsevier}
}

@article{chi2025diffusion,
  title={Diffusion policy: Visuomotor policy learning via action diffusion},
  author={Chi, Cheng and Xu, Zhenjia and Feng, Siyuan and Cousineau, Eric and Du, Yilun and Burchfiel, Benjamin and Tedrake, Russ and Song, Shuran},
  journal={The International Journal of Robotics Research},
  volume={44},
  number={10-11},
  pages={1684--1704},
  year={2025},
  publisher={Sage Publications Sage UK: London, England}
}

@article{zhao2023learning,
  title={Learning fine-grained bimanual manipulation with low-cost hardware},
  author={Zhao, Tony Z and Kumar, Vikash and Levine, Sergey and Finn, Chelsea},
  journal={arXiv preprint arXiv:2304.13705},
  year={2023}
}

@inproceedings{pelletier1994implementation,
  title={On the implementation and performance of impedance control on position controlled robots},
  author={Pelletier, Michel and Doyon, Michel},
  booktitle={Proceedings of the 1994 IEEE International Conference on Robotics and Automation},
  pages={1228--1233},
  year={1994},
  organization={IEEE}
}

@book{lynch2017modern,
  title={Modern robotics},
  author={Lynch, Kevin M and Park, Frank C},
  year={2017},
  publisher={Cambridge University Press}
}

@article{chi2024universal,
  title={Universal manipulation interface: In-the-wild robot teaching without in-the-wild robots},
  author={Chi, Cheng and Xu, Zhenjia and Pan, Chuer and Cousineau, Eric and Burchfiel, Benjamin and Feng, Siyuan and Tedrake, Russ and Song, Shuran},
  journal={arXiv preprint arXiv:2402.10329},
  year={2024}
}

@article{cadene2026lerobot,
  title={Lerobot: An open-source library for end-to-end robot learning},
  author={Cadene, Remi and Aliberts, Simon and Capuano, Francesco and Aractingi, Michel and Zouitine, Adil and Kooijmans, Pepijn and Choghari, Jade and Russi, Martino and Pascal, Caroline and Palma, Steven and others},
  journal={arXiv preprint arXiv:2602.22818},
  year={2026}
}

@article{butterworth1930theory,
  title={On the theory of filter amplifiers},
  author={Butterworth, Stephen and others},
  journal={Wireless Engineer},
  volume={7},
  number={6},
  pages={536--541},
  year={1930}
}

\newpage
\appendix
\section*{Appendix}

\section{Implementation Details}
\label{sec:implementation_details}

\subsection{Details for Network Architecture}
\label{appendix:a1_network_detail}

\paragraph{CompVLA.}
The model architecture used in our experiments is shown in Figure~\ref{fig:CompVLA}. Our implementation is based on the $\pi_0$ implementation in LeRobot v0.4.3~\cite{cadene2026lerobot}. Following the original $\pi_0$ design~\cite{black2024pi_0}, we use PaliGemma as the VLM backbone and a Gemma-based model for the action expert. We add a compliance expert with the same architecture as the action expert.

The action expert predicts the reference TCP twists and gripper commands over the motion chunk horizon $T=25$. It takes the robot state, flow time, and noisy action as input. It then applies blockwise causal attention to the key-value representations produced by the VLM and outputs a denoising vector. During inference, denoising starts from noise and runs for 10 steps. The final TCP twists are integrated from the current TCP pose to obtain the reference trajectory.

The compliance expert predicts the diagonal stiffness and virtual displacement values over the same horizon. It uses the VLM context and a TCP trajectory condition. Let $X_{t+i}$, $i=0,\ldots,T-1$, denote the TCP poses along the conditioning trajectory expressed in the base frame, with $X_t$ being the current TCP pose. We express this trajectory relative to $X_t$ and combine it with the corresponding gripper commands to define
\begin{equation}
Z_t = (\bar{X}_{t:t+T-1}, g_{t:t+T-1}),
\qquad
\bar{X}_{t+i} = X_t^{-1}X_{t+i}.
\end{equation}
The first relative pose is therefore $\bar{X}_t=I$. For each timestep, we flatten the relative rotation matrix and concatenate it with the relative position and gripper command to form a 13-dimensional condition. This condition is inserted into the unused padding channels of the noisy compliance input and embedded together with the flow time. The compliance expert then applies blockwise causal attention to the VLM key-value representations, following the same structure as the action expert, and outputs the denoising vector for stiffness and virtual displacement. During training, the TCP trajectory and gripper commands used to construct $Z_t$ are obtained from the dataset; during inference, they are obtained from the action expert's predictions. Together, the predicted TCP twists, gripper commands, stiffness values, and virtual displacements instantiate the motion chunk $M_t$ in Eq.~\eqref{eq:motion_chunk}.

\paragraph{Training.}
We train the VLM backbone, action expert, and compliance expert jointly with flow matching. Let $Y_t^a$ denote the target sequence of TCP twists and gripper commands over the horizon $T$, and let $Y_t^c$ denote the target sequence of diagonal stiffness and virtual displacement values over the same horizon. For a target sequence $Y_t$ at environment timestep $t$, we sample Gaussian noise $\epsilon$ and flow time $\tau$, and define the noisy target as
\begin{equation}
\tilde{Y}_t(\tau) = \tau \epsilon + (1-\tau)Y_t,
\end{equation}
where $\tau=1$ corresponds to noise and $\tau=0$ corresponds to the data target. Since inference integrates from $\tau=1$ to $\tau=0$, the target vector field is $\epsilon - Y_t$.

The action expert is trained with the conditional flow matching loss
\begin{equation}
\mathcal{L}_a = \mathbb{E}_{\tau \sim \mathcal{U}(0,1), \, Y_t^a \sim p(Y_t^a \mid O_t, l), \, \epsilon^a \sim \mathcal{N}(0, I)} \left[ \left\| v_\theta^a(\tilde{Y}_t^a(\tau), \tau \mid O_t, l) - (\epsilon^a - Y_t^a) \right\|_2^2 \right],
\end{equation}
where $Y_t^a = (\mathcal{V}_{t:t+T-1}, g_{t:t+T-1})$, $O_t$ represents the multimodal observation, and $l$ denotes the language instruction. The compliance expert is trained with
\begin{equation}
\mathcal{L}_c = \mathbb{E}_{\tau \sim \mathcal{U}(0,1), \, Y_t^c \sim p(Y_t^c \mid O_t, l, Z_t), \, \epsilon^c \sim \mathcal{N}(0, I)} \left[ \left\| v_\theta^c(\tilde{Y}_t^c(\tau), \tau \mid O_t, l, Z_t) - (\epsilon^c - Y_t^c) \right\|_2^2 \right],
\end{equation}
where $Y_t^c = (k_{t:t+T-1}, d_{t:t+T-1})$, and $Z_t$ is the conditioning sequence of ground-truth relative TCP poses and gripper commands constructed from the dataset. The final objective is
\begin{equation}
\mathcal{L} = \mathcal{L}_a + \mathcal{L}_c.
\end{equation}

Inputs and training targets are normalized by the preprocessor, and predicted outputs are unnormalized by the postprocessor during inference. We initialize the model weights from the \textit{pi0\_base} checkpoint provided through Hugging Face\footnote{\url{https://huggingface.co/lerobot/pi0_base}}. The compliance expert is initialized by copying the weights of the action expert. To better preserve the information in the pretrained checkpoint, we include joint angles as auxiliary prediction targets during training and discard their predictions at inference. These auxiliary channels are omitted from the notation above. We train CompVLA for 100,000 steps with a batch size of 32 in \texttt{bfloat16} precision, leveraging gradient checkpointing to ensure memory efficiency. Optimization is performed using the AdamW optimizer with a peak learning rate of $2.5 \times 10^{-5}$, which undergoes a warmup phase of 1,000 steps, decays to a minimum learning rate of $2.5 \times 10^{-6}$ over the subsequent 30,000 steps, and remains constant for the remainder of the training.

\subsection{Details for Control}
\label{appendix:a2_control_detail}
This section details the practical implementation of the robot control law. For notational simplicity, the time subscript $t$ is omitted throughout the following derivations. We denote the current TCP pose by $X=(R,p)$. First, the virtual target pose $X^v = (R^v, p^v) \in \mathrm{SE}(3)$ is established by deforming the reference pose $X_d=(R_d, p_d) \in \mathrm{SE}(3)$ using the predicted virtual displacement vector $d = [d^R; d^p] \in \mathbb{R}^6$,
\begin{equation}
    R^v = R_d \operatorname{Exp}(d^R), \quad p^v = p_d + R_d d^p,
\end{equation}
where $\operatorname{Exp}(\cdot)$ denotes the exponential map from $\mathbb{R}^3$ to $\mathrm{SO}(3)$. Since the virtual displacement components $d^R$ and $d^p$ are defined entirely within the reference TCP frame, the translational displacement is mapped to the base frame via the reference rotation matrix $R_d$.

As outlined in Section~\ref{subsec:learning_compliance}, distinct direction-dependent compliance frames are identified for both orientation and position tracking. The orientations of these compliance frames relative to the TCP frame are denoted by the rotation matrices $\tilde{R}^R$ and $\tilde{R}^p$, respectively. The diagonal stiffness parameters $k^R$ and $k^p$ provided by CompVLA are defined within these respective frames. To control the robot in the task space, the stiffness matrices relative to the TCP frame are computed for orientation and position as $K_d^R = \tilde{R}^R \text{diag}(k^R) (\tilde{R}^R)^\top$ and $K_d^p = \tilde{R}^p \text{diag}(k^p) (\tilde{R}^p)^\top$, respectively. These are then represented as a single $6\times6$ block diagonal matrix
\begin{equation}
    K_d = \begin{bmatrix} K_d^R & \mathbf{0} \\ \mathbf{0} & K_d^p \end{bmatrix} = \begin{bmatrix} \tilde{R}^R \text{diag}(k^R) (\tilde{R}^R)^\top & \mathbf{0} \\ \mathbf{0} & \tilde{R}^p \text{diag}(k^p) (\tilde{R}^p)^\top \end{bmatrix}.
\end{equation}
For damping, the diagonal parameters $c^R$ and $c^p$ follow the critical-damping heuristic $c_i = 2\sqrt{k_i}$, which assumes unit effective inertia. The task-space damping matrix $D_d$ is then formulated through an identical block-wise transformation of these diagonal matrices, yielding $D_d^R = \tilde{R}^R \text{diag}(c^R) (\tilde{R}^R)^\top$ and $D_d^p = \tilde{R}^p \text{diag}(c^p) (\tilde{R}^p)^\top$, which form the final $6\times6$ block diagonal matrix
\begin{equation}
    D_d = \begin{bmatrix} D_d^R & \mathbf{0} \\ \mathbf{0} & D_d^p \end{bmatrix} = \begin{bmatrix} \tilde{R}^R \text{diag}(c^R) (\tilde{R}^R)^\top & \mathbf{0} \\ \mathbf{0} & \tilde{R}^p \text{diag}(c^p) (\tilde{R}^p)^\top \end{bmatrix}.
\end{equation}
Based on the obtained $K_d$ and $D_d$, the robot is controlled by the generalized spring-damper relationship. The resulting task-space wrench $\mathcal{F}$ within the TCP frame is then computed as
\begin{equation}
    \mathcal{F} = -K_d X_{e} - D_d \mathcal{V}_{e},
\end{equation}
where $\mathcal{V}_{e}$ denotes the task-space twist error defined analogously to \eqref{eq:se3-error} relative to the virtual target, and $X_{e} = [R_e; p_e] \in \mathbb{R}^6$ represents the decoupled pose error evaluated in the TCP frame.
Specifically, the rotational error is defined as $R_e = \operatorname{Log}((R^{v})^\top R) \in \mathbb{R}^3$, and the translational error is given by $p_e = R^\top (p - p^{v}) \in \mathbb{R}^3$, where $\operatorname{Log}(\cdot)$ denotes the logarithmic map from $\mathrm{SO}(3)$ to $\mathbb{R}^3$. Like~\eqref{eq:se3-error}, this pose error does not depend on the choice of the global reference frame.

Finally, the task-space impedance wrench is mapped to the joint torque contribution $\tau_{\mathrm{imp}}$ through the body Jacobian $J(\theta)$,
\begin{equation}
    \tau_{\mathrm{imp}} = J^\top(\theta) \mathcal{F},
\end{equation}
where $\theta$ denotes the current joint configuration vector. This term implements the stiffness and damping contributions in~\eqref{eq:imp-law} with $\Lambda_{d}=\Lambda(\theta)$. Consequently, the anisotropic stiffness matrices establish the desired directional compliance, while tracking the virtual target pose implicitly regulates the contact forces, achieving the desired force modulation.

Compared with GIC, the implementation omits the feedforward terms $\Lambda(\theta)\,\dot{\mathcal{V}}_{d}^{\ast} + \mu(\theta, \mathcal{V})\,\mathcal{V}_{d}^{\ast}$, relies on the robot's low-level controller for gravity compensation, and uses the decoupled pose error above instead of~\eqref{eq:se3-error}. The first two simplifications follow common practice in impedance control, which often drops the mass compensation term and replaces the nonlinear dynamics compensation with gravity compensation when velocities are small~\cite{lynch2017modern}. Because of these simplifications and the time-varying stiffness, the stability result of~\citet{seo2024comparison} motivates this controller rather than guaranteeing its stability.

\section{Dataset Generation}

\subsection{Data Collection Setup}
\label{appendix:b1_data_collection_setup}

\begin{wrapfigure}{r}{0.48\textwidth}
    \centering
    \vspace{-13pt}
    \includegraphics[width=1\linewidth]{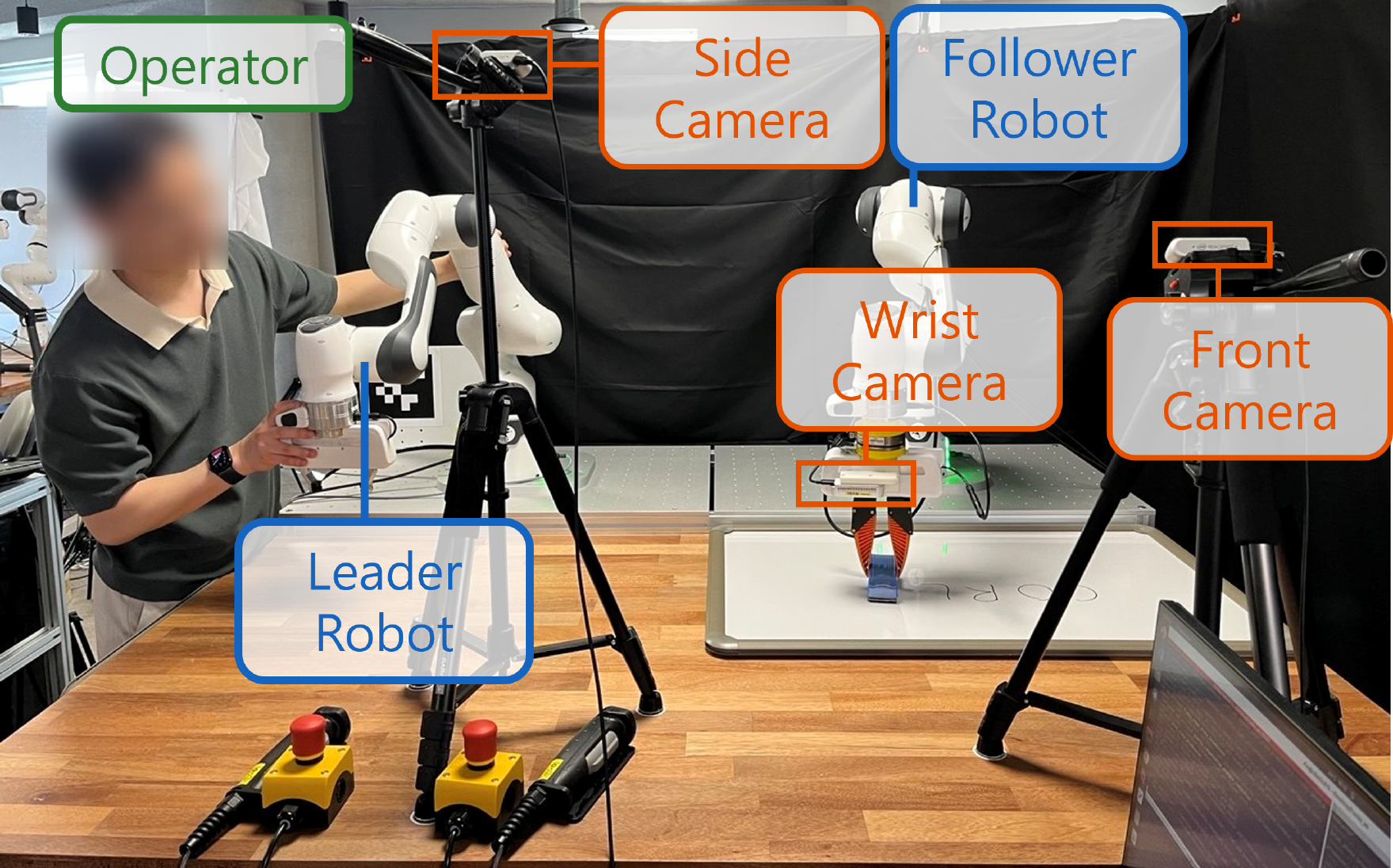}
    \vspace{-13pt}
    \caption{Teleop Data Collection Setup.}
    \vspace{-12pt}
    \label{fig:supp_teleop}
\end{wrapfigure}

Our data collection setup is illustrated in Figure~\ref{fig:supp_teleop}. The platform consists of two 7-DoF Franka Research 3 (FR3) arms, with the left arm acting as the leader and the right arm as the follower. The operator teleoperates the arm by moving the leader robot. The follower robot mimics the joint positions of the leader robot with slight compliance. To reduce the difficulty of teleoperation, haptic feedback is given to the leader robot so the operator can respond to the contact experienced by the follower robot. Figure~\ref{fig:supp_teleop} shows the operator teleoperating the system to collect data for the Erase Whiteboard task.

\subsection{Learning K from Demonstration}
\label{appendix:b2_learning_k}

To extract the time-varying diagonal stiffness $k_t = [k_t^R; k_t^p]$ from the human demonstrations, we provide the concrete formulation of the force-dependent heuristic outlined in Section~\ref{subsec:learning_compliance}. While classical variable impedance learning typically estimates stiffness based on electromyography (EMG) signals~\cite{ajoudani2012tele} or demonstration variance~\cite{calinon2010learning, kronander2013learning}, our approach, similar to ACP~\cite{hou2025adaptive}, dynamically modulates the stiffness along the axis of the applied force or torque based on its filtered magnitude at each timestep $t$. Specifically, the time-varying stiffness $k_{\text{low}}(|f_t|)$ is computed via piecewise linear interpolation according to the following formulation:
\begin{equation}
    k_{\text{low}}(|f_t|) = \begin{cases}
    k_{\text{max}}, & |f_t| < f_{\text{min}} \\
    k_{\text{max}} - (k_{\text{max}} - k_{\text{min}}) \frac{|f_t| - f_{\text{min}}}{f_{\text{max}} - f_{\text{min}}}, & f_{\text{min}} \le |f_t| \le f_{\text{max}} \\
    k_{\text{min}}, & |f_t| > f_{\text{max}}
    \end{cases}
\end{equation}
where $|f_t|$ denotes the magnitude of the filtered interaction force (for translation) or torque (for rotation) at time $t$. These four bounding parameters are configured independently for each space; specifically, we set $(k_{\text{max}}, k_{\text{min}}, f_{\text{min}}, f_{\text{max}})$ to $(500\,\text{N/m}, 200\,\text{N/m}, 1\,\text{N}, 15\,\text{N})$ for the translational space and $(30\,\text{Nm/rad}, 10\,\text{Nm/rad}, 0.1\,\text{Nm}, 2.5\,\text{Nm})$ for the rotational space. This yields eight hyperparameters in total, successfully accommodating the distinct physical scales of forces and torques across different tasks. Finally, the diagonal stiffness vectors $k_t^p$ and $k_t^R$ are constructed by assigning the modulated $k_{\text{low}}(|f_t|)$ to the first axis of each compliance frame, aligned with the corresponding force or torque direction, while the two orthogonal axes are maintained at $k_{\text{max}}$, thereby fully determining the stiffness profile $k_t = [k_t^R; k_t^p]$.

\subsection{Dataset Conversion from Raw Data}
\label{appendix:b3_dataset_conversion}

During human demonstrations, we record multiple modalities from the follower robot at their respective sampling rates: joint angles ($100$\,Hz), force/torque (F/T) measurements ($100$\,Hz), gripper widths ($15$\,Hz), and three camera views ($30$\,fps). To align these asynchronous streams, all modalities are synchronized using global timestamps and downsampled to a uniform frequency of $10$\,Hz. For continuous states, including joint angles, gripper widths, and F/T signals, synchronization is achieved via time-aware linear interpolation. For the video streams, downsampling is performed by selecting the frame closest to each synchronized timestamp. Utilizing the synchronized joint states, we compute the Tool Center Point (TCP) poses and twists via forward kinematics.

The F/T signals undergo a multi-stage preprocessing pipeline to ensure signal quality. First, they are processed with gravity and inertial force compensation. Subsequently, the measurements are transformed from the F/T sensor frame to the TCP frame using a wrench transformation based on the adjoint matrix $\text{Ad}(T_{\text{S}}^{\text{TCP}})$, where $T_{\text{S}}^{\text{TCP}} \in \mathrm{SE}(3)$ denotes the homogeneous transformation matrix representing the spatial calibration offset from the sensor frame to the TCP frame. To ensure smoothness, the transformed signals are filtered using a Butterworth low-pass filter~\cite{butterworth1930theory}, followed by magnitude clipping to limit excessively large force or torque targets. Finally, after the comprehensive dataset statistics have been compiled, the processed streams are converted into the LeRobot~\cite{cadene2026lerobot} format and saved for policy training.

\section{Additional Experimental Results and Visualizations}
\label{sec:aexperiments}

\begin{figure}[tbp]
    \centering
    \vspace{-10pt}
    \includegraphics[width=1.0\linewidth]{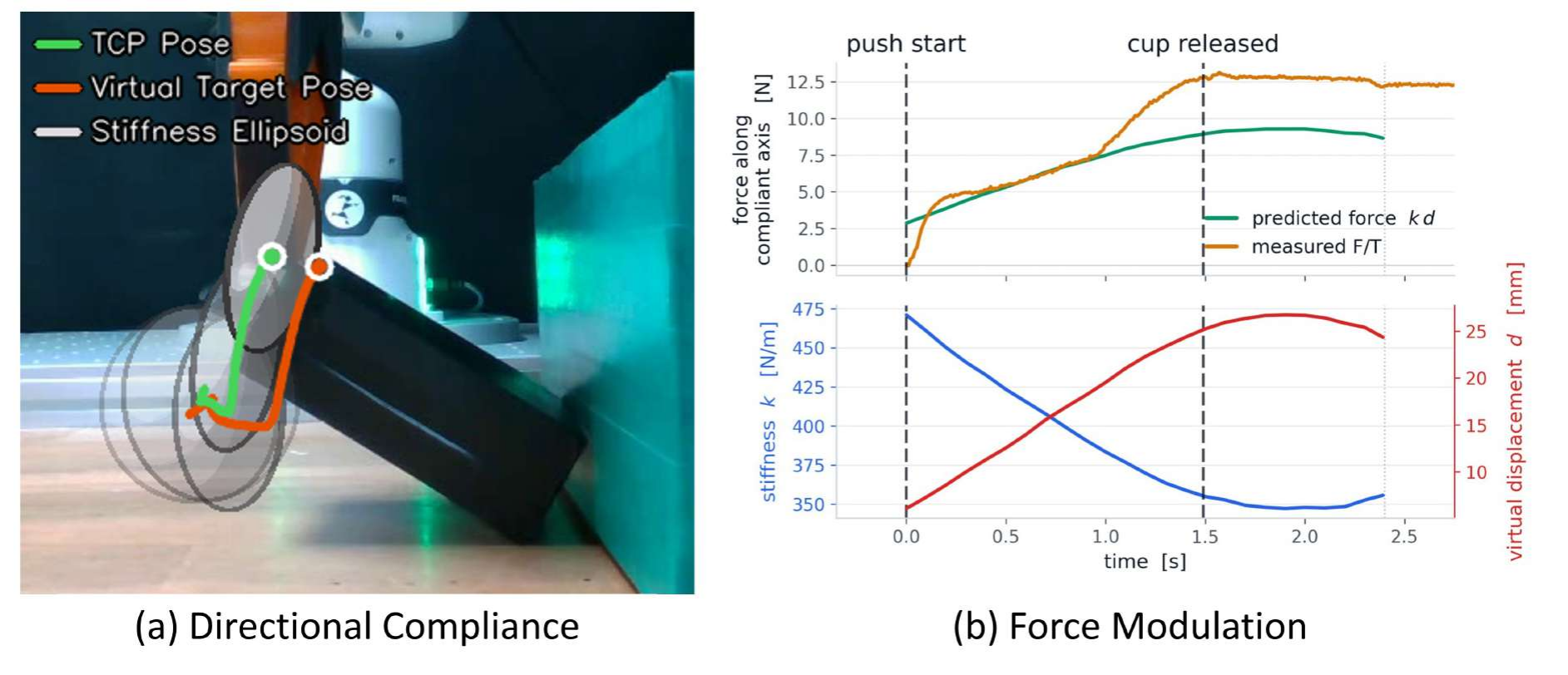}
    \caption{Visualization of compliance predictions. (a) TCP pose, virtual target pose, and stiffness ellipsoid during Pivot. The ellipsoid is short along the direction toward the pivot center and elongated along the tangential direction. (b) Predicted stiffness and virtual displacement, their force reference $kd$, and measured contact force during Cup Dispense. The measured contact force shows that CompVLA regulates the pressing force to the predicted force level $kd$. Together, (a) and (b) illustrate the two core capabilities of CompVLA, Directional Compliance and Force Modulation.}
    \label{fig:compliance_analysis}
\end{figure}

\subsection{Analyzing Performance Bottlenecks in Cup Dispense}
\label{appendix:c1_contact_analysis}

\begin{wraptable}{r}{0.5\textwidth}
    \vspace{-7pt}
    \centering
    \caption{Cup Dispense stage-wise success rates (\%), averaged over two runs of 20 trials per stage.}
    \label{tab:cup_stages}
    \vspace{-5pt}
    \begin{tabular}{@{}lcc@{}}
        \toprule
        \textbf{Method} & \textbf{Reaching} & \textbf{Pushing} \\
        \midrule
        $\pi_0$ & \textbf{52.5} & 67.5 \\
        CompVLA (ours) & 45.0 & \textbf{92.5} \\
        \bottomrule
    \end{tabular}
    \vspace{-7pt}
\end{wraptable}

To investigate which stage limits performance on Cup Dispense, we conduct additional evaluations of reaching and pushing. Pushing starts from the reached pose, or from near the button if reaching fails. CompVLA achieves slightly lower reaching performance than $\pi_0$ (45.0\% vs. 52.5\%), while improving pushing success from 67.5\% to 92.5\% (Table~\ref{tab:cup_stages}). This improvement highlights the benefit of learned compliance for reliable execution of the contact stage.

\subsection{Visualizing Compliance and Task Execution}
\label{appendix:c2_visualization}

\begin{figure}[ht]
    \centering
    \rotatebox{90}{%
        \begin{minipage}{0.8\textheight}
            \centering
            \includegraphics[width=\linewidth]{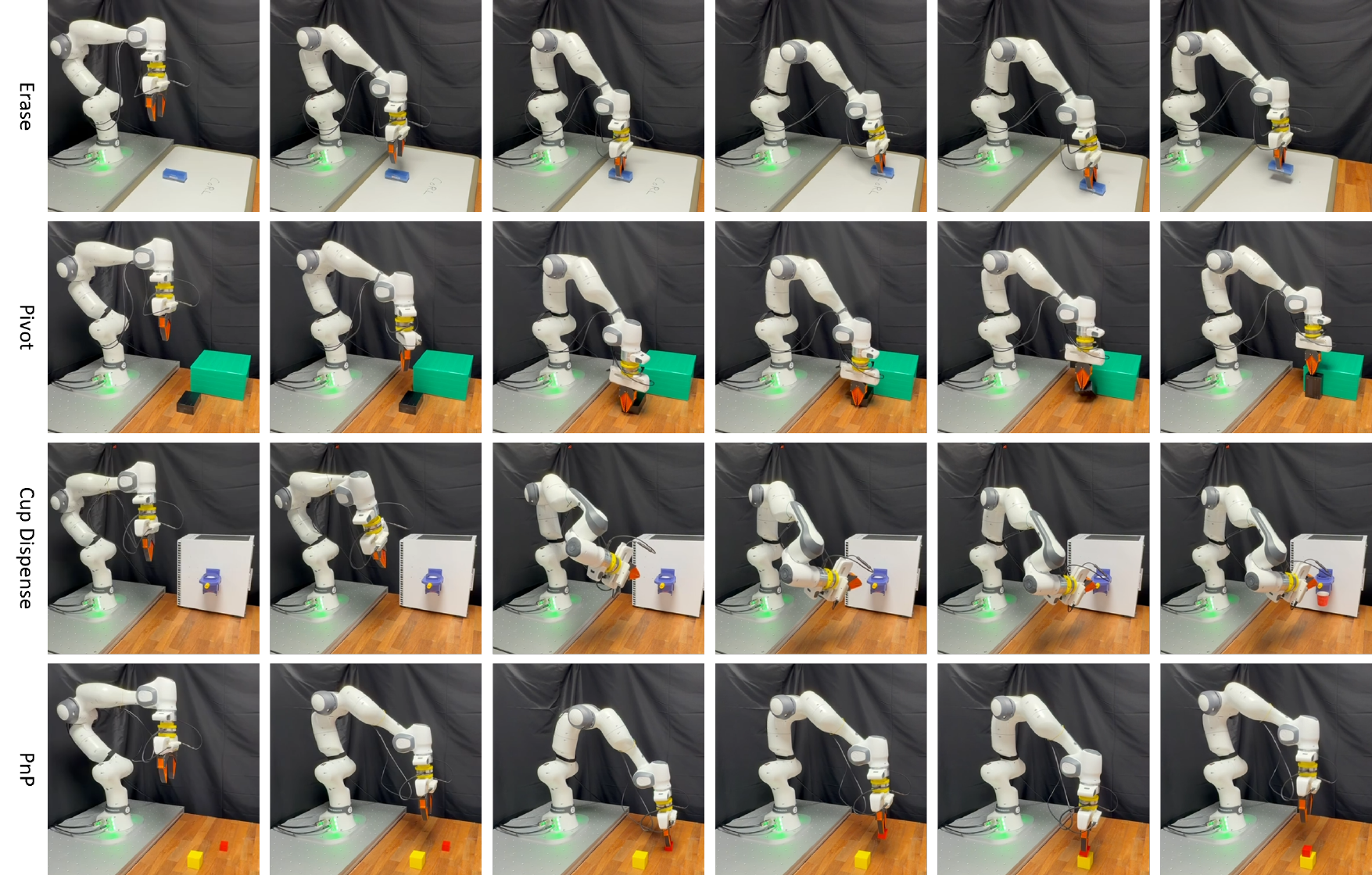}
            \captionof{figure}{Key frames of tasks used for training and evaluation.}
            \label{fig:sup_task_vis}
        \end{minipage}%
    }
\end{figure}

In this subsection, we visualize CompVLA's compliance predictions and measured contact forces to illustrate its directional compliance and force modulation during interaction (Figure~\ref{fig:compliance_analysis}). We also present key frames of the experimental tasks to illustrate the detailed motion throughout task execution (Figure~\ref{fig:sup_task_vis}).

\end{document}